\documentclass[conference]{IEEEtran}

\ifCLASSINFOpdf
  
\else
  
\fi

\usepackage{graphicx}
\usepackage{amsmath}
\usepackage{amsfonts}
\usepackage{algorithm}
\usepackage{algorithmic}
\usepackage{url}

\begin{document}
\title{Enhancing Precision in Tactile Internet-Enabled Remote Robotic Surgery: Kalman Filter Approach }


\author{
\IEEEauthorblockN{Muhammad Hanif Lashari}
\IEEEauthorblockA{Department of Electrical and\\Computer Engineering\\
Iowa State University\\
Ames, Iowa, USA.\\
Email: mhanif@iastate.edu}
\and
\IEEEauthorblockN{Wafa Batayneh}
\IEEEauthorblockA{Department of Electrical and\\Computer Engineering\\
Iowa State University\\
Ames, Iowa, USA.\\
Email: batayneh@iastate.edu}
\and
\IEEEauthorblockN{Ashfaq Khokhar}
\IEEEauthorblockA{Department of Electrical and\\Computer Engineering\\
Iowa State University\\
Ames, Iowa, USA.\\
Email: ashfaq@iastate.edu}

}

\maketitle

\begin{abstract}

Accurately estimating the position of a patient's side robotic arm in real time in a remote surgery task is a significant challenge, particularly in Tactile Internet (TI) environments. This paper presents a Kalman Filter (KF) based computationally efficient position estimation method. The study also assume no prior knowledge of the dynamic system model of the robotic arm system. Instead, The JIGSAW dataset, which is a comprehensive collection of robotic surgical data, and the Master Tool Manipulator's (MTM) input are utilized to learn the system model using System Identification (SI) toolkit available in Matlab. We further investigate the effectiveness of KF to determine the position of the Patient Side Manipulator (PSM) under simulated network conditions that include delays, jitter, and packet loss. These conditions reflect the typical challenges encountered in real-world Tactile Internet applications. The results of the study highlight KF's resilience and effectiveness in achieving accurate state estimation despite network-induced uncertainties with over 90\% estimation accuracy.


\end{abstract}

\IEEEpeerreviewmaketitle

\section{Introduction}
The Tactile Internet (TI) is a cutting-edge concept that is part of the next generation of mobile communication systems, known as 6G. Super-fast and reliable networks will enable delivery of skills and touch-based communication, leading to major societal changes. Unlike the regular internet, TI promises to offer seamless global connectivity, thanks to its use of advanced 6G technology. There will be different ways to interact with digital technology in the future \cite{1}.

TI, driven by groundbreaking technological advancements, focuses on real-time transmission of touch using state-of-the-art haptic equipment and robotics. This innovation heralds a shift from mere content delivery to a dynamic system of skill-set exchange over the Internet. It promises an ultra-responsive and ultra-reliable network connectivity, which is crucial for applications where real-time control and feedback are imperative \cite{2}. 

Central to the TI’s ambitious goal is its stringent network performance requirements. For mission-critical applications, TI necessitates a network latency typically ranging between 1-10 milliseconds and a remarkably high packet delivery ratio of 99.99999\%. These specifications are vital due to the sensitivity of human touch and the potentially catastrophic outcomes of any failure in these systems \cite{3}.

TI's applications, notably in domains like remote surgery, demand not only ultra-low latency but also a high level of reliability and security. The variability in latency requirements, often less than 10 milliseconds, is dictated by the specific nature and dynamicity of the application. TI aspires to achieve an ultra-low end-to-end round-trip latency of 1 millisecond, setting a new benchmark in network performance \cite{4}.

Diverging from traditional robotic surgery, TI enables remote robotic surgery where a surgeon operates on a distant patient through a network. This requires an unprecedented level of transparency, ensuring that the surgeon’s actions are accurately mirrored in the patient-side domain and inversely surgeon is precisely aware of the robotic arm position at the patient side. Achieving this bidirectional awareness in the face of communication-induced delays is a significant technical challenge \cite{5}\cite{6}.

Moreover, the current 5G mobile networks only partially meet these stringent requirements of TI. Issues such as delay, packet loss, and jitter can critically impact the stability and safety of remote robotic surgery systems \cite{7}\cite{8}. Due to the extremely time sensitive requirements of the application domain, it is important to explore computationally lightweight solutions. This paper introduces the application of Kalman Filter --assisted by an offline System Identification learning module--as a solution to these challenges. By accurately estimating the position of the PSM arm, the KF enhances the reliability and precision of TI applications, particularly in the high-stakes realm of remote surgery.

\section{Related Work}
In \cite{9}, the authors address packet loss and delay challenges in remote robotic surgery within a 5G Tactile Internet environment, advocating for a Gaussian process regression (GPR) approach to predict and compensate for delayed/lost messages. Two kernel versions of the sequential randomized low-rank and sparse matrix factorization method (1-SRLSMF and SRLSMF) were introduced to scale GPR for handling delayed/lost data in training datasets. However, this approach faces challenges due to the computational complexity of Gaussian processes, especially kernel matrix inversion, which escalates with increasing data points. 

In \cite{10}, the authors proposed a method based on deep learning and Convolutional Neural Networks (CNN) for evaluating surgical skills in robot-assisted surgery. It introduces a deep learning framework to assess skills by mapping motion kinematics data to skill levels using a Deep CNN. The study also highlights limitations of CNN, including the need for improved labeling methods, optimization of the deep architecture, and exploring ways to visualize deep hierarchical representations to uncover hidden skill patterns.

The data is collected through the da Vinci Research Kit (dVRK), which is a specialized set of robotic tools for testing surgical procedures. It is the very first generation model of the da Vinci Surgical System (dVSS) by a company called Intuitive Surgical \cite{11}. This kit has been thoroughly studied by experts in order to understand system dynamics \cite{12} \cite{13}. This research delves into the PSM arm movements, using JIGSAW data of the surgical system to learn its mechanics.

The main objective of this study is explore the efficacy of Kalman Filter to estimate the position of the PSM's arm, even when the network experiences delays, jitter, or data packet loss. These issues are a significant challenge, especially when bidirectional touch information and precise control are necessary across the network. We thoroughly evaluate the performance of the proposed system by simulating different network conditions. The robotic arm positions are simulated using the dVRK module.

\section{Methodology}
The KF is a well signal processing algorithm that employs efficient, recursive computation for process state estimation, aiming to minimize the mean squared error. It supports estimations of past, present, and future states, even under system uncertainties. It was introduced by Rudolf E. Kalman in 1960 \cite{urrea2021kalman}, and it has been effectively used in fields requiring accurate and real-time estimation, such as navigation. It is particularly effective in systems where data is uncertain or noisy, which is common in remote robotic surgery stemming from sampling noise and network uncertainties.

\begin{figure}[htbp]
    \centering
    \includegraphics[width=0.5\textwidth]{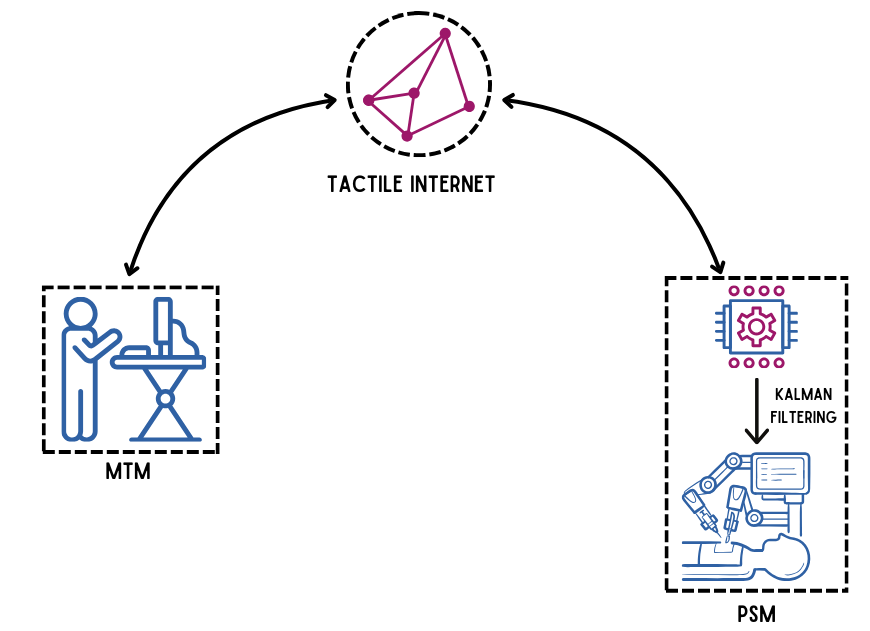}
    \caption{Remote Robotic Surgery Framework Utilizing TI and KF for Enhanced PSM Precision}
    \label{model}
\end{figure}

\subsection{System Model}

The proposed architecture \cite{9} as shown in Fig.\ref{model}, is a remote robotic surgery system facilitated by Tactile Internet. The system is compartmentalized into three primary domains: the surgeon-side domain, the patient-side domain, and the network domain, each playing an integral role in the surgical procedure's execution.

\textbf{Surgeon-Side Domain}: The surgeon-side domain is comprised of an ergonomically designed surgeon console/master tool manipulator (MTM) and the operating surgeon. The surgeon interacts with the console, which in turn captures the surgeon's gestures and translates them into haptic commands. These commands encapsulate the surgeon's intended surgical maneuvers, encompassing aspects such as force, orientation, and kinematic parameters.

\textbf{Patient-Side Domain}: The patient-side domain hosts the PSM and the patient. Upon reception of the haptic commands, the PSM, equipped with an estimation KF algorithm (in our case KF), interprets these inputs to estimate and enact the precise movements corresponding to the surgeon's inputs. The KF algorithm is pivotal for real-time estimation and correction of the robot's arm position, as it assists in maintaining the fidelity of the surgical gestures amidst potential perturbations in signal transmission.

\textbf{Network Domain}: Central to the communication bridge between the surgeon and patient domains, the network domain is tasked with delivering low-latency and ultra-reliable connectivity.

\begin{figure*}[htbp]
    \centering
    \includegraphics[width=\textwidth]{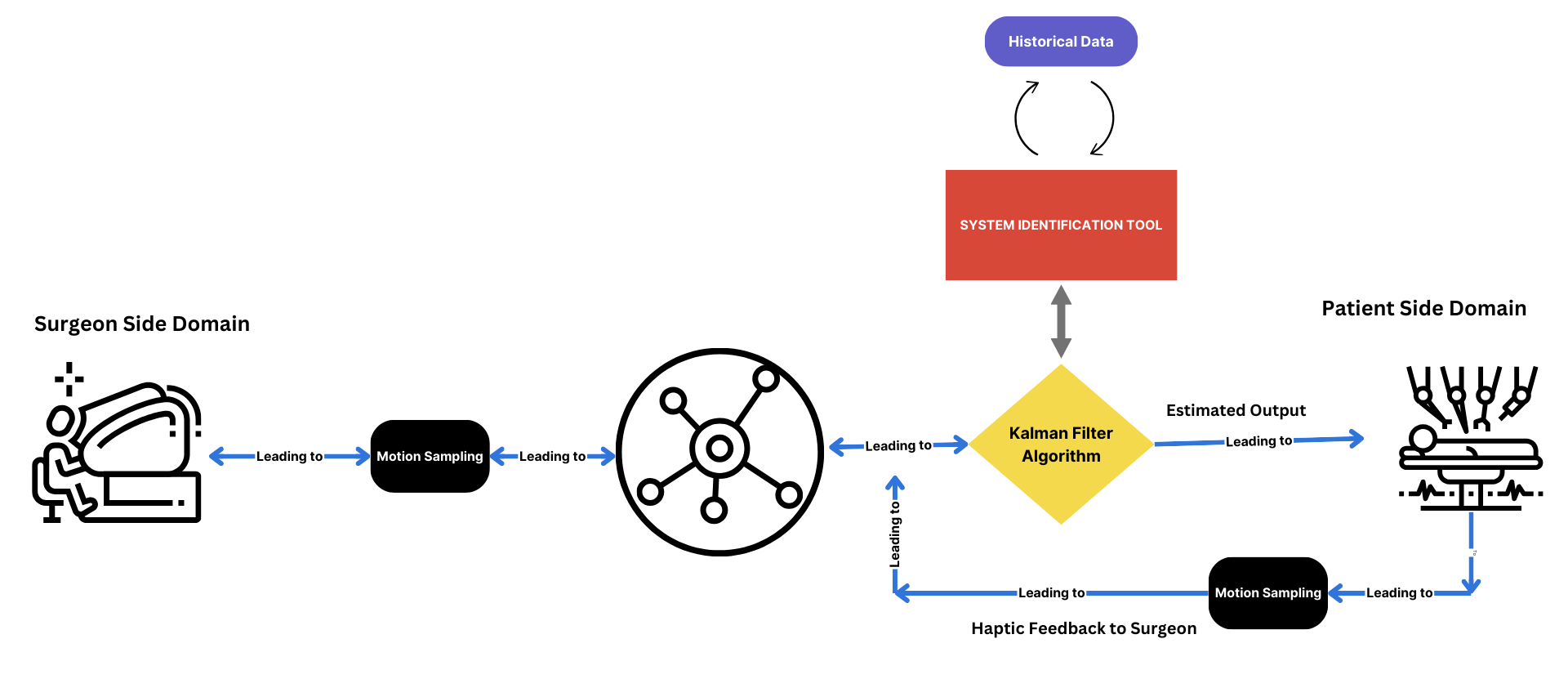}
    \caption{Operational Sequence of Remote Surgery Using TI and KF Algorithm}
    \label{fig:model1}
\end{figure*}

\textbf{Operational Workflow}:
The block diagram, as shown in Fig.\ref{fig:model1}, illustrates the operational sequence that initiates with the surgeon interacting with the MTM, generating a set of haptic commands. These commands are transmitted through the network domain, leveraging TI technology. The movements are processed using a KF Algorithm, which utilizes historical data for system identification and generates an accurate estimation of the required movement. This estimated output is then executed by the PSM in the Patient Side Domain. Feedback from the PSM is sent back to the surgeon, providing vital tactile information to inform the surgeon's subsequent movements. This feedback loop is essential for the precision and safety of remote surgical procedures.

\subsection{Kalman Filter Implementation for State Estimation with Tactile Internet Network Effects}
For the sake of completeness, in this section we briefly describe the Kalman Filter fundamentals. The KF has numerous applications in technology. A common application is the guidance, navigation, and control of vehicles, especially aircrafts and exploration robots \cite{urrea2021kalman}. It is also widely used in signal processing and Quantum Systems \cite{ma2022review}.

The system is represented as follows:

\begin{itemize}
    \item \textbf{State Equation (System Dynamics):} The true state of the system evolves according to the discrete-time state-space model but is affected by the network characteristics before being observed at the PSM side.
    \begin{equation}
        \tilde{x}_k = A\tilde{x}_{k-1} + Bu_{k-1} + w_{k-1}
    \end{equation}
    where:
    \begin{itemize}
        \item \( \tilde{x}_k \) is the predicted state vector of the system at time \( k \) before accounting for network effects.
        \item \( A \) is the state transition matrix that models the system dynamics from one timestep to the next.
        \item \( B \) is the control input matrix that translates the input commands into changes in the state.
        \item \( u_{k-1} \) is the control input vector applied at the previous time step \( k-1 \).
        \item \( w_{k-1} \) represents the process noise at the previous time step, which encompasses the inherent uncertainties in the system model as well as additional perturbations such as those introduced by the network jitter and packet loss.
    \end{itemize}
    
    \item \textbf{Network Effects:} The network introduces additional deviations to the state vector as it is transmitted from the MTM to the PSM. These deviations are modeled as:
    \begin{equation}
        x_k = \tilde{x}_k + n_{d} + n_{j} + n_{p}
    \end{equation}
    where:
    \begin{itemize}
        \item \( x_k \) is the state vector as it arrives at the PSM, having been affected by the network.
        \item \( n_{d} \) models the deviation caused by network delay, which can vary depending on the current network conditions.
        \item \( n_{j} \) models the deviation caused by network jitter, representing the variability in the delay.
        \item \( n_{p} \) models the deviation caused by packet loss, which can result in intermittent losses of information.
    \end{itemize}
    
    \item \textbf{Measurement Equation (Observation Model):} The PSM observes the system state with its own sensors, which can be represented by the measurement equation.
    \begin{equation}
        z_k = Hx_k + v_k
    \end{equation}
    where:
    \begin{itemize}
        \item \( z_k \) is the measurement vector at time \( k \).
        \item \( H \) is the observation matrix that relates the state vector to the measurements.
        \item \( v_k \) represents the measurement noise at time \( k \), reflecting the sensor noise and other observational inaccuracies.
    \end{itemize}
\end{itemize}

Given these models, the KF operates in two steps, Prediction and Update, to estimate the system state:

\begin{itemize}
    \item \textbf{Prediction Step:} The filter predicts the state of the system at the next time step along with the estimation uncertainty.
    \begin{align}
        \hat{x}_k^- &= A\hat{x}_{k-1} + Bu_{k-1} \\
        P_k^- &= AP_{k-1}A^T + Q
    \end{align}
    where:
    \begin{itemize}
        \item \( \hat{x}_k^- \) is the a priori estimate of the state vector before the measurement at time \( k \) is taken into account.
        \item \( P_k^- \) is the a priori estimate of the state covariance, indicating the uncertainty of the prediction.
        \item \( Q \) is the covariance matrix of the process noise, quantifying the expected variance in the predictions due to the inherent uncertainty in the system dynamics and the effect of the network.
    \end{itemize}
    
\item \textbf{Update Step:} The filter then incorporates the new measurement to refine its estimate of the state vector and update the estimation uncertainty.
    \begin{align}
        K_k &= P_k^- H^T (HP_k^- H^T + R)^{-1} \\
        \hat{x}_k &= \hat{x}_k^- + K_k(z_k - H\hat{x}_k^-) \\
        P_k &= (I - K_kH)P_k^-
    \end{align}
    where:
    \begin{itemize}
        \item \( K_k \) is the Kalman gain at time \( k \), which determines how much the predictions should be adjusted based on the new measurement.
        \item \( \hat{x}_k \) is the a posteriori estimate of the state vector after incorporating the measurement at time \( k \).
        \item \( P_k \) is the a posteriori estimate of the state covariance, indicating the updated uncertainty of the state estimate.
        \item \( R \) is the covariance matrix of the measurement noise, quantifying the expected variance in the measurements.
        \item \( I \) is the identity matrix, with the same dimensions as \( P_k^- \).
    \end{itemize}
\end{itemize}

The KF uses these equations to continuously estimate the system's state in the presence of noise and uncertainties, including those introduced by the Tactile Internet. The filter's ability to estimate the true state in such an environment is a measure of its robustness and effectiveness.

\section{Dataset and Experimental Setup}
\subsubsection{\textbf{Source and Composition of the JIGSAWS Dataset}}
\begin{figure*}[htbp]
  \centering
  \includegraphics[width=\textwidth]{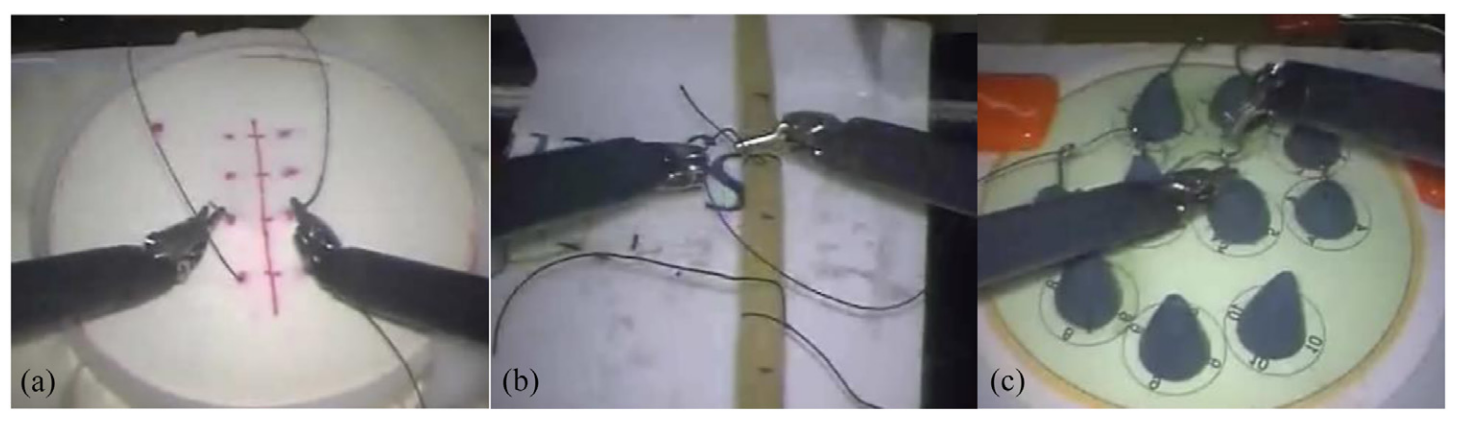}
  \caption{Three surgical tasks: (a) Suturing, (b) Knot-tying and (c) Needle-passing  \cite{anh2020towards}.}
  \label{fig:data}
\end{figure*}

In order to evaluate the remote robotic surgery system, the JIGSAWS dataset has been used, which is a comprehensive surgical skill dataset developed by the Computational Interaction and Robotics Laboratory at Johns Hopkins University \cite{DaVinciSurgicalSystem2022}. JIGSAWS stands for the JHU-ISI Gesture and Skill Assessment Working Set and encompasses kinematic, video, and gesture data from three elementary surgical tasks performed using the da Vinci surgical robot: suturing, knot tying, and needle passing as shown in Fig.\ref{fig:data}.

\begin{figure*}[htbp]
    \centering
    \includegraphics[width=\textwidth]{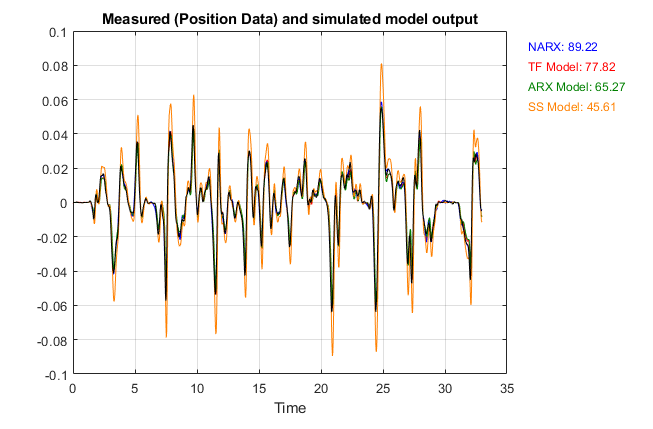}
    \caption{Matlab SIT Model Identification}
    \label{fig:model}
\end{figure*}

\begin{figure*}[htbp]
    \centering
    \includegraphics[width=\textwidth]{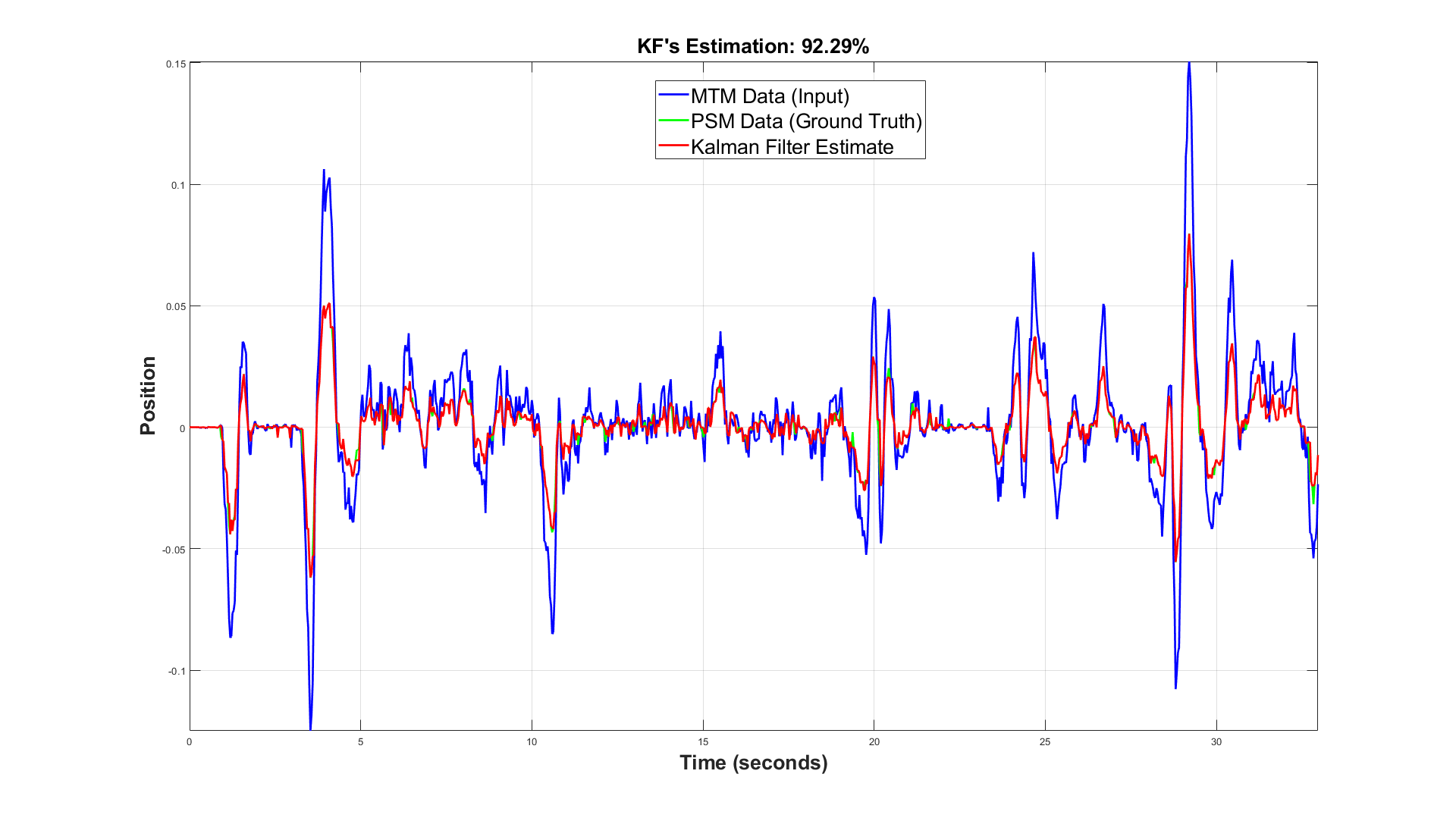}
    \caption{Comparison of PSM data with the KF estimated output under simulated network conditions. The estimation percentage reflects the accuracy of the KF in tracking the state of the PSM}
    \label{fig:Output}
\end{figure*}

The dataset captures the motion data of both the master controls and the corresponding slave manipulators (end-effectors) during the execution of the tasks. A diverse group of eight participants, ranging from novices to experts, contributed to the dataset. Each participant was tasked with performing each surgical task five times, resulting in a total of 40 trials. The kinematic data is comprised of 76 motion variables recorded at a sampling frequency of 30 Hz \cite{lefor2020motion}.

\subsubsection{\textbf{System Identification with MATLAB SIT}}
While implementing the KF in the da Vinci Surgical System
(dVSS), accurate modeling of system dynamics and noise characteristics is paramount. However, the proprietary nature of dVSS poses a challenge as its dynamics and noise characteristics are not publicly disclosed \cite{7}.

In \cite{wang2019convex}, authors present a Convex Optimization-based Dynamic Model Identification Package for the da Vinci Research Kit (dVRK) - a teleoperated surgical robotic system. The package is designed to model the mechanical components of the dVRK and identify dynamic parameters subject to physical consistency. It addresses the need for accurate dynamic models before implementing robust model-based control algorithms and is open-source, making it feasible for use on similar robots.
In \cite{ferro2022coppeliasim}, authors describe the development of a dynamic simulator for the dVRK (PSM) in the CoppeliaSim robotic simulation environment. The simulator aims to accurately predict the behavior of the real robot by integrating the kinematic and dynamic properties, including the double parallelogram and the counterweight mechanism.

The JIGSAWS dataset presents a high-dimensional challenge, with 76 inputs reflecting the complex motions of the MTM and PSM. Given the inherent nonlinearity in the kinematic behavior of the robotic system, particularly in the x, y, and z positional data of the PSM arm, a nonlinear modeling approach was deemed necessary. This study utilized the nonlinear Auto-regressive with exogenous input (ARX) model in the MATLAB System Identification Tool (SIT), which is well-suited for capturing the intricate dynamics of linear and nonlinear systems.

The nonlinear ARX model allowed us to model the nonlinear relationships between the numerous inputs and the three-dimensional position of the PSM arm. This model structure was favored because of its ability to approximate the nonlinearities present in the surgical robot's movements without necessitating an exhaustive modeling of the system's physics.

Through systematic experimentation within the SIT environment, this work evaluated the nonlinear ARX model's capacity to faithfully represent the PSM's kinematics. The model's parameters were iteratively adjusted to optimize the prediction of the PSM arm's position, ensuring that the resulting model could serve as a reliable foundation for the KF's estimations.

In addition to the NARX model, this study explored alternative modeling frameworks such as Transfer Function models, State-Space models, and ARX models to ensure a comprehensive evaluation.
Figure~ref{fig:model} displays the implementation and comparison of four different models: NARX, TF, SS, and ARX.
A best-fit ratio accompanies the graphical results, indicating each model's ability to predict the system's output. The right side of the figure shows these ratios, clearly delineating the superior performance of the NARX model in capturing the PSM arm's positional behavior.

\subsubsection{\textbf{Cross-Validation of Model Predictions}}

To validate the nonlinear ARX model, we used a cross-validation methodology. We employed an independent dataset, which was different from the one used for system identification, to evaluate the model's predictive performance. Our main goal was to assess how well the model could predict the PSM arm's x, y, and z positions. These positions are crucial for accurately translating haptic commands during real-time surgery.

Additionally, this paper presents an efficient algorithm that uses a KF approach to address network-related challenges in remote robotic surgery. It accurately estimates the PSM position, taking into account network delay, jitter, and packet loss, as detailed in Algorithm 1.

\begin{algorithm}
\caption{KF with Network Effects for State Estimation}
\begin{algorithmic}[1]

\STATE \textbf{Input:} MTM, PSM, A, B, C, Q, R
\STATE \textbf{Output:} z\_est, MSE, Est\%

\STATE Init: $x\_est, P, prev\_y, N, dt, nd, nj, np$
\FOR {$k = 2$ to $N$ Samples}
    \STATE $del\_k \gets \max(1, k - \text{round}(nd / dt + \text{randn}() \times nj / dt))$
    \STATE $y \gets (\text{rand}() > np) ? PSM[del\_k] : prev\_y$
    \STATE $prev\_y \gets y$, $x\_pred \gets A x\_est[k-1] + B MTM[k-1]$
    \STATE $P\_pred \gets A P A^T + Q$
    \FOR {$d = 1$ to $\text{size}(C, 1)$}
        \STATE $K \gets P\_pred C[d]^T / (C[d] P\_pred C[d]^T + R[d])$
        \STATE $x\_est[k] \gets x\_pred + K (y[d] - C[d] x\_pred)$
        \STATE $P \gets (I - K C[d]) P\_pred$
    \ENDFOR
\ENDFOR
\STATE $z\_est \gets C x\_est$, Calc MSE, Est\%

\end{algorithmic}
\end{algorithm}

\section{Results}

The implementation of the KF for state estimation through a network characterized by delay, jitter, and packet loss was evaluated. The network parameters were set to simulate a best-effort network scenario, reflecting conditions that might commonly be encountered in real-world Tactile Internet applications.

The network simulation parameters were as follows:
\begin{itemize}
    \item Network delay (\( n_d \) in ms): This represents a constant time delay that every packet experiences during transmission over the network. In \cite{varga20205g}, TI delay range mentioned for 5G services and use-cases.
    \item Jitter variance (\( n_j \) ms):  This is the variance of the jitter, indicating the degree of random fluctuation in the timing of packet arrivals around the mean network delay. In \cite{malekzadeh2023performance}, discusses the impact of jitter in 5G networks on the performance of real-time services, values of less than 0.01 seconds are associated with good performance. 
    \item Packet loss probability (\( n_p \) in \%): This is the likelihood that any given packet will be lost during transmission and not reach its destination. Values between 0.01 and 0.1 (1\% to 10\%) are often used in simulations to study the impact of packet loss \cite{fettweis2014tactile}.
\end{itemize}

Under these conditions, the KF estimated the PSM's state using MTM inputs. The estimation's effectiveness was measured by the percentage accuracy, reflecting the match between the estimated and actual PSM states. 

Figure \ref{fig:Output} shows the comparison between MTM input data, observed PSM data, and KF estimated output under simulated network conditions. The achieved estimation accuracy of 83.47\% demonstrates the KF's effectiveness in maintaining high accuracy despite network uncertainties, as detailed in Table. \ref{tab:results}.

\begin{table}[h]
    \centering
    \small
    \caption{Results of KF Estimation with Different Network Conditions}
    \label{tab:results}
    \begin{tabular}{|p{1.5cm}|p{1.5cm}|p{1.5cm}|p{1.5cm}|}
        \hline
        Jitter Variance (ms) & Network Delay (ms) & Packet Loss Probability & KF Estimation in \% \\
        \hline
        0 & 0 & 0.00 & 99.78 \\
        2 & 5 & 0.10 & 92.29 \\
        5 & 7 & 0.20 & 83.47 \\
        6 & 3 & 0.18 & 86.90 \\
        4 & 8 & 0.13 & 88.86 \\
        4 & 5 & 0.20 & 84.17 \\
        6 & 5 & 0.15 & 86.26 \\
        \hline
    \end{tabular}
\end{table}

\begin{itemize}
    \item \textbf{Optimal Conditions:} Achieves peak accuracy of 99.78\% with no network impairments, ideal for remote surgery.
    \item \textbf{Mild Network Impairments:} Slight accuracy drop to 92.29\% under mild impairments (jitter 2 ms, delay 5 ms, packet loss 10\%), still reliable.
    \item \textbf{Moderate to High Impairments:} Accuracy varies from 83.47\% to 88.86\% with higher impairments (jitter up to 6 ms, delay up to 8 ms, packet loss up to 20\%), indicating reduced but operational performance.
    \item \textbf{Worst-Case Scenario:} Lowest performance at 83.47\% under severe conditions (jitter 5 ms, delay 7 ms, packet loss 20\%), highlighting the need for robust algorithms.
\end{itemize}

\section{Conclusions and Future Research Directions}
Robotic surgery is carried out in dynamic environments that come with uncertainties. To ensure safe surgical operations, the KF's design is tailored to account for such uncertainties, providing reliable estimations. This paper highlights the effectiveness of KF in remote robotic surgery scenarios, especially within the Tactile Internet context. By leveraging the JIGSAWS dataset for system identification and simulation, the study has successfully demonstrated the KF's capability to accurately estimate the PSM's position despite significant network challenges such as delay, jitter, and packet loss. This contribution  marks a significant step forward in ensuring precision and reliability in tele-operative surgical procedures. It also lays a foundation for future research in this field, aiming to enhance the safety and efficacy of remote surgery in network-constrained environments.

This research could be further extended by optimizing the performance of the KF for even more complex and challenging network conditions. This could be achieved by integrating adaptive filtering techniques that can better handle dynamic network changes. Furthermore, we plan to explore the integration of lightweight machine learning algorithms with KF to enhance the prediction accuracy and adaptability in complex surgical scenarios.


\section*{Acknowledgment}
The work in this paper was partially supported by the Palmer Department Chair and the Richardson Professorship Endowments.

\bibliographystyle{unsrt}
\bibliography{reference}

\end{document}